\pdfoutput=1

\documentclass[11pt]{article}

\usepackage[]{naacl2021}

\usepackage{times}
\usepackage{latexsym}

\usepackage{graphicx}
\usepackage{subfigure}
\usepackage{caption}
\usepackage{algorithm} 
\usepackage{mathrsfs}
\usepackage{algpseudocode}  
\usepackage{amsthm}
\usepackage{amsfonts}
\usepackage{amssymb}
\usepackage{enumitem}

\usepackage[utf8]{inputenc}

\usepackage{microtype}

\title{Few-Shot Domain Adaptation for Grammatical Error Correction \\ via Meta-Learning }

\author{Shengsheng Zhang\textsuperscript{1,2},
	Yaping Huang\textsuperscript{1}, 
	Yun Chen\textsuperscript{3},
	Liner Yang\textsuperscript{2},\\
	\textbf{Chencheng Wang\textsuperscript{4},
	Erhong Yang\textsuperscript{2}}
	\\
	\textsuperscript{1}{Beijing Jiaotong University, Beijing, China}
	\\
	\textsuperscript{2}{Beijing Language and Culture University, Beijing, China}
	\\
	\textsuperscript{3}{Shanghai University of Finance and Economics, Shanghai, China}
	\\
	\textsuperscript{4}{Beijing University of Technology, Beijing, China}
	\\
}

\begin{document}
\maketitle
\begin{abstract}
Most existing Grammatical Error Correction (GEC) methods based on sequence-to-sequence mainly focus on how to generate more pseudo data to obtain better performance.  Few work addresses few-shot GEC domain adaptation. In this paper, we treat different GEC domains as different GEC tasks and propose to extend meta-learning to few-shot GEC domain adaptation without using any pseudo data. We exploit a set of data-rich source domains to learn the initialization of model parameters that facilitates fast adaptation on new resource-poor target domains. We adapt GEC model to the first language (L1) of the second language learner. To evaluate the proposed method, we use nine L1s as source domains and five L1s as target domains. Experiment results on the L1 GEC domain adaptation dataset demonstrate that the proposed approach outperforms the multi-task transfer learning baseline by 0.50 $F_{0.5}$ score on average and enables us to effectively adapt to a new L1 domain with only 200 parallel sentences. 
\end{abstract}

\section{Introduction}\label{Intr}
Grammatical Error Correction (GEC) aims to correct errors in text. For example, ``He notice the picture.'' can be corrected to “He notices the picture.". A GEC system takes an incorrect sentence as input and outputs the corresponding correct sentence. With the development of deep learning, GEC has drawn the attention of many researchers during the last few years.

Most existing methods \citep{DBLP:journals/corr/abs-1801-08831,junczys-dowmunt-etal-2018-approaching,zhao-etal-2019-improving} frame GEC as a sequence-to-sequence (seq2seq) task and have obtained high performance on the general domain while using a large number of training examples. However, these seq2seq-based models cannot gain satisfactory performance in special GEC domains due to domain shift and the limited in-domain data. For instance, \citet{nadejde-tetreault-2019-personalizing} use the GEC model trained on the general domain to test on specific domains and find that the performance drops dramatically. 
One way to tackle this issue is transfer learning \citep{nadejde-tetreault-2019-personalizing}, in which a GEC model is pretrained on the high-resource general domain and then fine-tuned on a low-resource target domain. Although leading to empirical improvements in the target domain, this method suffers from model over-fitting and catastrophic forgetting when the in-domain data is insufficient~\cite{Sharaf2020MetaLearningFF}.
\begin{figure}[t]
	\begin{center}
		{ 
			\includegraphics[width=0.4\textwidth]{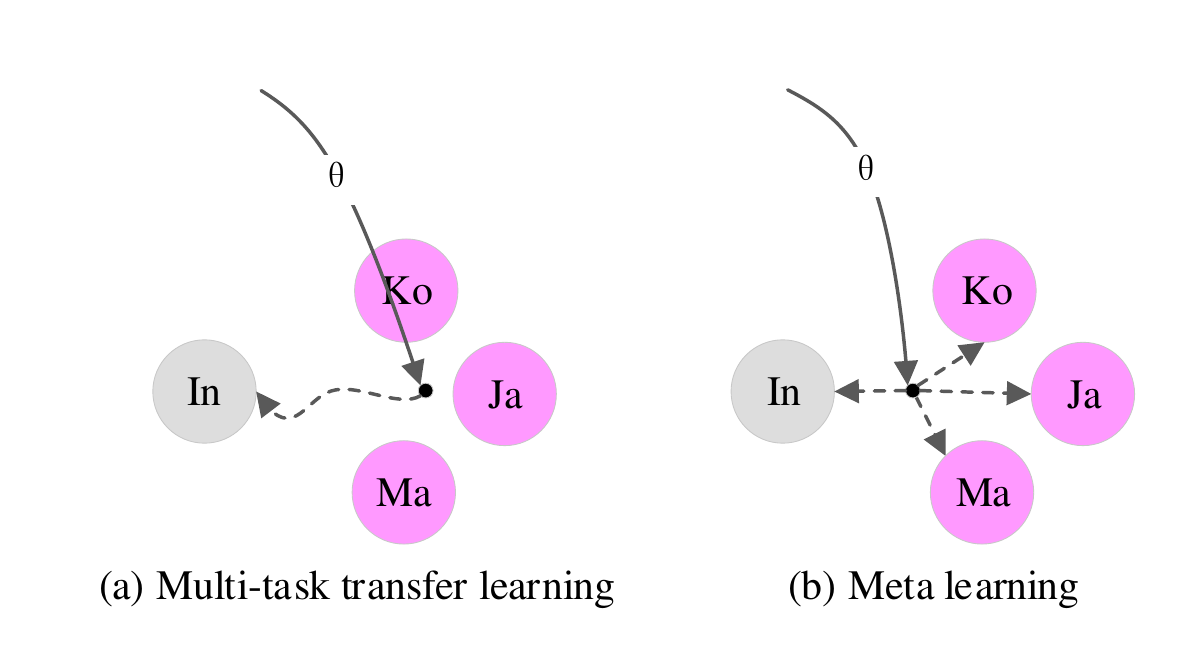}
		}
	\end{center}
	\caption{Difference between multi-task transfer learning and
		meta learning. Solid lines denote the learning of initial parameters and dashed lines are the path of fine-tuning. Pink and gray represent the source task and target task, respectively. }
	\label{difference}
\end{figure}
In this paper, we frame GEC system for different domains as different tasks and propose a meta-learning method for few-shot GEC domain adaptation.  
Specifically, we use model-agnostic meta-learning
algorithm (MAML; \citet{ DBLP:journals/corr/FinnAL17}) to learn  the initialization of model parameters from high-resource domains, which can quickly adapt to a new target domain with a minimal amount of data. Fig.\ref{difference} shows the difference between our method and the multi-task transfer learning method in~\citet{nadejde-tetreault-2019-personalizing}. Their method first trains GEC model on multi-domain data and then fine-tunes it on a target domain. 

To evaluate the proposed method, we adapt GEC model to Chinese as a Second Language (CSL) learner's first language (L1). We construct a few-shot GEC domain adaptation dataset by making use of 4 resource-poor L1s as the test domains and the rest 10 L1s as the source and valid domains. Our experiments on the constructed dataset show that our method can effectively adapt to a new domain using only 200 parallel sentences and outperform the multi-task transfer learning method by 0.50 $F_{0.5}$ score on average. To our best knowledge, we are the first to apply meta-learning to GEC.

\section{Method}
\subsection{GEC Domain Adaptation}\label{different task}
Given an erroneous sentence $ X=\{x_1,...,x_M\}$ and a learner's domain $d$, a Neural Machine Translation (NMT)-based model for domain-aware GEC models the conditional probability of the output sentence $Y=\{y_1,...,y_N\}$ with neural networks as follows:
\begin{equation}
p(Y|X,d;\theta) = \prod_{t=1}^{N}p(y_t|y_{1:t-1},x_{1:M},d;\theta),
\label{eq.1}
\end{equation}
where $\theta$ is a set of model parameters. Following \citet{madotto-etal-2019-personalizing}, we first adapt $\theta$ to the learner's domain $d$ and then model the output sentence conditional on the erroneous input sentence with:
\begin{equation}
p(Y|X;\theta_d) = \prod_{t=1}^{N}p(y_t|y_{1:t-1},x_{1:M};\theta_d),
\label{eq.2}
\end{equation}
where $\theta_d$ is the set of domain-aware model parameters. A learner's domain can be defined with different criterion, such as the L1 and the proficiency level. In this paper, we use the L1 as the criterion and adapt a GEC system to the learner's L1. Since our method is agnostic to the definition of domains, it can be easily extended to other type of domain-aware GEC systems.
\subsection{Few-Shot GEC Domain Adaptation via Meta Learning}
We propose to apply the model-agnostic meta-learning (MAML; \citet{DBLP:journals/corr/FinnAL17} in few-shot GEC domain adaptation.  
We use MAML to learn a good initialization of model parameters $\theta^0$, which can quickly adapt to new domains using few training examples. We call the proposed meta-learning method for GEC domain adaptation as MetaGEC. 

We define a set of source tasks $\mathscr{T}=\{\mathcal{T}_{d_1},...,\mathcal{T}_{d_k}\}$, where each task $\mathcal{T}_{d_i}$ is a GEC system of a specific domain $d_i$ and $k$ is the number of learner's domains. For each meta-learning episode, we randomly sample a task $\mathcal{T}_{d_i}$ from $\mathscr{T}$. Then we sample two batches independently from task $\mathcal{T}_{d_i}$'s data, a support batch $D_{d_i}^s$ and a query batch $D_{d_i}^q$. We first use $D_{d_i}^s$ to update the GEC model parameters $\theta$ as follows:

\begin{equation}
\theta^{'}_{d_i}=\theta-\alpha\nabla_{\theta}\mathcal{L}_{D_{d_i}^s}(\theta) ,
\label{eq.3}
\end{equation}
where $\alpha$ is the learning rate and $\mathcal{L}$ is the cross-entropy loss function:
\begin{equation}
\mathcal{L}_{D_{d_i}^s}(\theta)=-\sum_{D_{d_i}^s}\log p(Y|X,\theta).
\label{eq.loss}
\end{equation}
After that, we evaluate the updated parameters $\theta_{d_i}^{'}$ on $D_{d_i}^q$ and update the original model parameters $\theta$ with gradient computed from this evaluation. It is possible to aggregate multiple episodes of source tasks before updating $\theta$. Therefore the original model parameters $\theta$ are updated as follows: 
\begin{equation}
\theta=\theta-\beta\sum_{d_i}\nabla_{\theta}\mathcal{L}_{D_{d_i}^{q}}(\theta_{d_i}^{'}) ,
\label{eq.4}
\end{equation}
where $\beta$ is the meta learning rate. The full algorithm is shown in Algorithm \ref{Algorithm 1}. 
\begin{algorithm}[t]
	\caption{Meta learning for few-shot GEC domain adaptation}
	\label{Algorithm 1}
	\textbf{Require:} $\mathscr{T}$: set of source tasks \\
	\textbf{Require:} $\alpha,\beta$: step size hyperparameters
	\begin{algorithmic}[1]
		\State Randomly initialize $\theta$
		\While{not done}{}
		\State Sample batch of tasks $\mathcal{T}_{d_i} \sim \mathscr{T}$
		\For{\textbf{all }$\mathcal{T}_{d_i}$}
		\State $(D_{d_i}^s,D_{d_i}^q) \sim D_{d_i}$
		\State Evaluate$\nabla_{\theta}\mathcal{L}_{D_{d_i}^s}(\theta)$ using $D_{d_i}^s$
		\State Compute adapted parameters with gra-
		\Statex \qquad \quad dient descent:   $\theta_{d_i}^{'}=\theta-\alpha\nabla_{\theta}\mathcal{L}_{D_{d_i}^s}(\theta)$
		\EndFor
		\State Update meta $\theta:$ 
		\Statex   \quad $\;$    $\theta=\theta-\beta\sum_{d_i}\nabla_{\theta}\mathcal{L}_{D_{d_i}^q}(\theta_{d_i}')$
		\EndWhile
	\end{algorithmic}
\end{algorithm}
The update of meta parameters involves second-order partial derivatives, which is computationally expensive. In our experiments, we use a first-order approximation to save memory consumption following previous work~\cite{gu-etal-2018-meta}. 
\begin{table}[t]
	\centering
	\begin{tabular}{lrrr}
		\hline
		Corpus & \#Sentence &\#SrcToken &\#TgtToken \\
		\hline
		Lang-8 & 1.09M & 14M & 15M \\
		HSK & 88K & 1.78M & 1.76M \\
		\hline
	\end{tabular}
	\caption{\label{corpustable}Data statistics for the Lang-8 and HSK datasets}
\end{table}
After the meta-training phrase, task-specific learning is done on a small amount of examples from a new target task $\mathcal{T}_d$, in order to obtain a task-specific model $\theta_d$.

\section{Experiments}
\subsection{Settings}
\noindent\textbf{Dataset}\quad  We use two datasets in our experiments: Lang-8\footnote{\url{https://lang-8.com}} and HSK\footnote{\url{http://hsk.blcu.edu.cn/}}. Both dataset are written by CSL learners and corrected by Chinese native speakers. We tokenize the datasets by jieba\footnote{\url{https://github.com/fxsjy/jieba}} and apply Byte Pair Encoding~\cite{sennrich-etal-2016-neural} to limit vocabulary size.\footnote{\url{https://github.com/rsennrich/subword-nmt}} We first pretrain our model on Lang-8 and then study GEC domain adaptation on the HSK dataset with the pretrained model. Table \ref{corpustable} shows the statistics of both datasets.
HSK consists of examination essays written by CSL learners with fourteen different L1s. First, we choose four domains with the least data as the test domains, including German (De),  Russian (Ru), French (Fr)  and  Mongolian (Mo). Then, we randomly sample one domain from the rest domains as the valid domain, while the other domains serve as the source domains. Specifically, we use Indonesian (In) as the valid domain, and Korean (Ko), Traditional Chinese (Zh-tw), Japanese (Ja), Singapore English (En-Sg), Malay (Ma), Burmese (Bu), Thai (Th), Vietnamese (Vi) and English (En) as the source domains. For each source domain, we sample 1000 parallel sentences as the in-domain dataset. For valid domain, we sample 200, 800, and 400 parallel sentences as the in-domain training set, development set, and test set respectively. For each test domain, we sample 200 parallel sentences as the in-domain training set, and divide the rest data in HSK into development set and test set according to a two-to-one ratio. The valid and test domains are also called target domains. We use the ERRANT\footnote{\url{https://github.com/chrisjbryant/errant}} to make the gold edits of grammatical errors in sentences of  each test set.

\noindent\textbf{GEC System}\quad\label{Implementation details}
\begin{table*}[t]
	\centering
	\begin{tabular}{l|c c c| c}
		\hline
		Target Task & \textit{No Fine-tuning} & \textit{Fine-tuning} & \textit{MTL+Fine-tuning} & \textit{MetaGEC} \\
		\hline
		In & 19.18 & 25.40 & 37.09 & \textbf{37.46} \\ 
		\hline
		De & 24.43 & 30.03 & 37.76 & \textbf{39.43} \\
		Ru & 21.44 & 33.98 & \textbf{40.15} & 39.14 \\
		Fr & 29.10 & 35.48 & 43.19 & \textbf{43.49} \\
		Mo & 29.30 & 36.72 & 48.07 & \textbf{49.21} \\
		\hline
		Average &24.69 & 32.32  & 41.25 & \textbf{41.75} \\
		\hline
	\end{tabular}
	\caption{
		\label{MetaGEC result} 
		$F_{0.5}$ score on the test set of the target tasks, where In is the valid task and all other L1s are the test tasks.}  
\end{table*}
We utilize the Transformer \citep{DBLP:journals/corr/VaswaniSPUJGKP17} implemented by fairseq\footnote{\url{https://github.com/pytorch/fairseq}} as our GEC model. We follow the model configure in \texttt{transformer\_wmt\_en\_de} and set batch size to 4000 tokens. For pretraining on Lang-8, we follow the training instructions in~\citet{ott2018scaling}.\footnote{\url{https://github.com/pytorch/fairseq/blob/v0.9.0/examples/scaling_nmt/README.md}} For meta training, we use the same Adam optimizer except that we set \texttt{lr=1e-5} for the outer loop and \texttt{lr=1e-7} for the inner loop. At test time, we fine-tune the model on the target task's 
training set with \texttt{lr=5e-4}.
For all models, we translate with beam search using \texttt{beam\_size=12}.

\noindent\textbf{Baselines}\quad We compare MetaGEC with three baselines:
(1) \textit{No Fine-tuning}~\cite{Sharaf2020MetaLearningFF}: the method that evaluates the pretrained GEC model on the target task's test set;
(2) \textit{Fine-tuning}~\cite{Sharaf2020MetaLearningFF}: the method that fine-tunes the pretrained GEC model on the target task's training data directly; (3) \textit{MTL+Fine-tuning}~\cite{nadejde-tetreault-2019-personalizing}: the multi-task transfer learning method we discussed in Section~\ref{Intr}. It first fine-tunes the pretrained GEC model on all data of the source tasks in a multi-task learning framework, and then fine-tunes the resulting model on the target task's training data.  As an evaluation metric, we use $F_{0.5}$ score computed by applying the MaxMatch\footnote{\url{https://www.comp.nus.edu.sg/~nlp/conll14st.html}} ($M^2$) scorer \citep{dahlmeier-ng-2012-better}.
We repeat the baselines and our method three times with different seeds and report the averaged score.
\subsection{Results}
Table \ref{MetaGEC result} shows the evaluation results of \textit{MetaGEC} and the baselines. Overall, \textit{MetaGEC} outperforms all the baselines, improving \textit{No Fine-tuning}, \textit{Fine-tuning} and \textit{MTL+Fine-tuning} by 17.06, 9.43 and 0.50 $F_{0.5}$ on average. This indicates that \textit{MetaGEC} has successfully found a good initialization of model parameters for fast domain adaptation. We also observe that for Ru, \textit{MetaGEC} performs worse than the baseline \textit{MTL+Fine-tuning}. Ru benefits the most when fine-tuning with in-domain data (\textit{No Fine-tuning} to \textit{Fine-tuning}) among all five target tasks. In contrast, it benefits the least from multi-task learning (\textit{Fine-tuning} to \textit{MTL+Fine-tuning}). We hypothesize that for Ru, fine-tuning with in-domain data is more important than the way we choose to utilize the data contained in source tasks. Since \textit{MetaGEC} is different from \textit{MTL+Fine-tuning} in the way of utilizing data from the source tasks, our hypothesis also partially explains the degraded performance of \textit{MetaGEC} on Ru.

\begin{figure}[t]
	\centering
	\subfigure{\includegraphics[width=5.7cm]{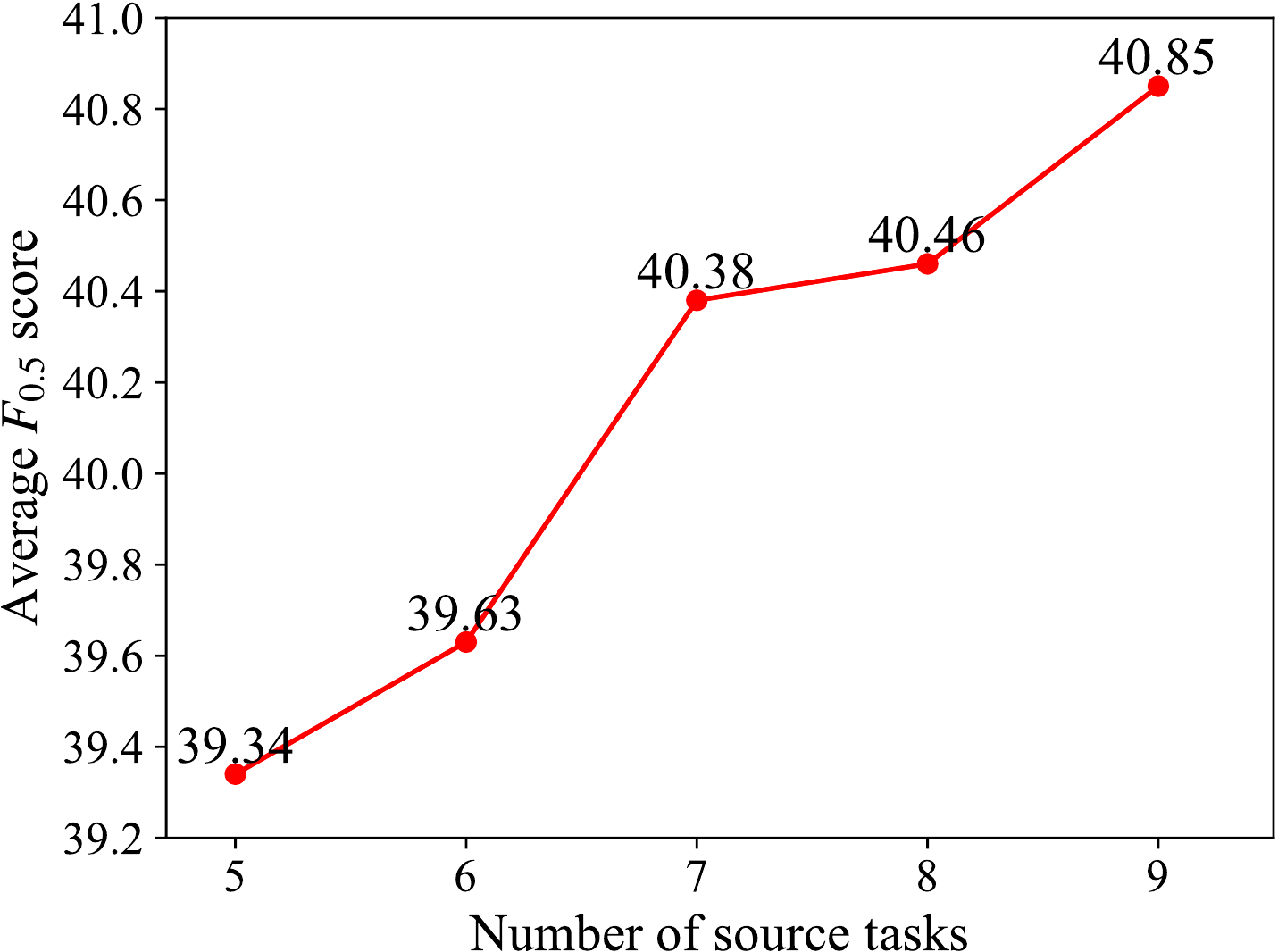}}
	\caption{Impact of the number of source tasks. We report the averaged $F_{0.5}$ score on the test sets of the five target tasks (In, De, Ru, Fr, Mo).}
	\label{Impact}
\end{figure}

To study the impact of the number of source tasks, we experiment with different number of source tasks and report the averaged $F_{0.5}$ score on the test sets of the five target tasks, as shown in Fig.~\ref{Impact}. Note that we only ran the experiments once here. We use Ko, Zh-tw, Ja, Ma and Bu as the source tasks when the number of source tasks is 5, and gradually add Th, En-Sg, En and Vi when increasing the number of source tasks from 5 to 9. We observe that when including more source tasks at the meta training phase, we can obtain better performance on the target tasks, demonstrating that better initialization model can be learned with more source tasks. 
\section{Related Work}
\noindent\textbf{Grammatical Error Correction}\quad The traditional GEC approaches include two categories: specific rule-based methods \citep{5387835, DBLP:journals/corr/cmp-lg-9607001} and statistical machine translation (SMT)-based approaches \citep{brockett-etal-2006-correcting,junczys-dowmunt-grundkiewicz-2014-amu}. Specific ruled-based methods only correct certain types of errors in the text. 
SMT-based approaches greatly improve the performance of GEC. But they are surpassed by deep learning-based methods. \citet{junczys-dowmunt-etal-2018-approaching} cast GEC as a low-resource NMT task.
Due to the limited public data, many works \citep{lichtarge-etal-2019-corpora,kiyono-etal-2019-empirical,wang2019controllable,kaneko2020encoderdecoder} pay attention to how to generate more pseudo data to improve the performance of neural GEC models.\\ 
\noindent\textbf{GEC Domain Adaptation}\quad
\citet{rozovskaya-roth-2011-algorithm} use Naive Bayes classifier to adapt a model to the L1 of the learner.
\citet{chollampatt-etal-2016-adapting} first train a neural network joint model on the data labeled by L1 of the learner and then integrate it into a SMT based GEC system.   
\citet{nadejde-tetreault-2019-personalizing} utilize transfer learning method to adapt a model to different domains.\\ 
\noindent\textbf{Meta Learning}\quad Recently, meta-learning \citep{Lake1332, DBLP:journals/corr/AndrychowiczDGH16, DBLP:journals/corr/FinnAL17} has attracted lots of attention. 
Meta-learning aims at solving how to achieve fast adaption on new data. Current meta-learning methods can be classified into two categories: 1)  Learning strategies and policies \citep{DBLP:journals/corr/AndrychowiczDGH16}. 2) Learning good initial parameters of model \citep{DBLP:journals/corr/FinnAL17}.  Many works have applied meta-learning to Natural Language Processing tasks, such as low-resource NMT \citep{gu-etal-2018-meta}, personalizing dialogue agents \citep{madotto-etal-2019-personalizing} and few-shot NMT adaptation \citep{Sharaf2020MetaLearningFF}.
\section{Conclusion }
In this paper, we introduce MetaGEC, a model-agnostic meta-learning algorithm for few-shot GEC domain adaptation. MetaGEC exploits a set of data-rich source domains to learn the initialization of model parameters that facilitates fast adaptation for a new target domain with a minimal amount of training examples. Experiment results demonstrate the effectiveness of the proposed method. In the future, we will apply different meta-learning methods in the GEC task.

%


\bibliographystyle{acl_natbib}
\bibliography{anthology,custom}
\appendix

%

\end{document}